\documentclass{article} 
\usepackage{iclr2027_conference,times}

\usepackage{amsmath,amsfonts,bm}

\def\eqref#1{equation~\ref{#1}}

\def\1{\bm{1}}

\DeclareMathAlphabet{\mathsfit}{\encodingdefault}{\sfdefault}{m}{sl}
\SetMathAlphabet{\mathsfit}{bold}{\encodingdefault}{\sfdefault}{bx}{n}

\usepackage[most]{tcolorbox}
\usepackage{hyperref}
\usepackage{listings}
\usepackage{url}
\usepackage[table]{xcolor}
\usepackage{booktabs}
\usepackage{subcaption}
\usepackage{multirow}
\usepackage{graphicx}
\usepackage{adjustbox}
\usepackage{xcolor}
\usepackage{fvextra}
\usepackage{wrapfig}
\definecolor{codebg}{HTML}{F6F8FA}
\definecolor{codefr}{HTML}{BFC7D1}
\lstdefinestyle{pycode}{
  language=Python,
  basicstyle=\ttfamily\footnotesize,
  backgroundcolor=\color{codebg},
  frame=single, framerule=0.7pt, rulecolor=\color{codefr},
  breaklines=true, columns=fullflexible, keepspaces=true,
  showstringspaces=false, commentstyle=\color{gray},
  xleftmargin=7pt, xrightmargin=7pt, aboveskip=8pt, belowskip=8pt,
}
\newcommand{\gain}[1]{\textsuperscript{\textcolor{blue}{\scriptsize $\uparrow$#1\%}}}
\lstdefinestyle{reasoningtrace}{
  basicstyle=\ttfamily\scriptsize,
  breaklines=true,
  breakatwhitespace=false,
  columns=fullflexible,
  keepspaces=true,
  showstringspaces=false,
  upquote=true,
  frame=single,
  rulecolor=\color{black!15},
  backgroundcolor=\color{black!1},
  xleftmargin=0.5em,
  xrightmargin=0.5em,
  aboveskip=0.4em,
  belowskip=0.6em,
  literate=
    {→}{{$\rightarrow$}}1
    {∈}{{$\in$}}1
    {≡}{{$\equiv$}}1
    {≤}{{$\leq$}}1
    {≥}{{$\geq$}}1
}
\iclrfinalcopy
\title{MicroMouse: Masking Harmful Supervision in Long CoT via Relative Potential}

\author{
Xiaowei Zhu\textsuperscript{1,2},
Hengzhu Tang\textsuperscript{3},
Zhenyu Zhang\textsuperscript{3}\thanks{Project lead.}, \\
\textbf{
Shuohuan Wang\textsuperscript{3},
Yubing Ren\textsuperscript{1,2}\thanks{Corresponding author.},
Yanan Cao\textsuperscript{1,2}\footnotemark[2]
}\\
        \textsuperscript{1}Institute of Information Engineering, Chinese Academy of Sciences, Beijing, China \\ 
        \textsuperscript{2}School of Cyber Security, University of Chinese Academy of Sciences, Beijing, China \\
        \textsuperscript{3}Baidu Inc.,  Beijing, China \\
        \texttt{\{zhuxiaowei, renyubing, caoyanan\}@iie.ac.cn} \\
        \texttt{\{tanghengzhu, zhangzhenyu07, wangshuohuan\}@baidu.com}}

\begin{document}

\maketitle
\lhead{Preprint. Under review.}

\maketitle
\begin{abstract}
Long chain-of-thought (CoT) trajectories have become an important source of supervision for training large reasoning models, yet conventional supervised fine-tuning (SFT) typically optimizes all tokens in these trajectories indiscriminately. However, even successful long CoTs may contain harmful supervision, motivating the need for methods that can automatically identify and mask such signals during SFT. Inspired by the \emph{MicroMouse} maze-solving process, we view a long CoT as a search trajectory from the input problem toward the target answer. This perspective motivates maintaining a goal-relative measure, analogous to a MicroMouse's spatial coordinates, to quantify how close the current reasoning state is to the target answer. Building on this perspective, we propose \textsc{MicroMouse}, an automated selective-SFT framework for masking harmful supervision in long CoTs. In \textsc{MicroMouse} we introduce a \emph{Relative Potential} (REP) to measure how closely each reasoning state supports the gold answer. Unlike the perplexity of the gold answer alone, REP measures relative support by contrasting the gold answer with a matched foil. This contrast cancels likelihood variations shared by the pair, yielding a more stable signal of progress toward the target answer. By tracking the REP trajectory, \textsc{MicroMouse} identifies consecutive potential decreases as erroneous exploration and post-peak reasoning as redundant wandering, and masks these harmful supervision during SFT. Extensive experiments across multiple student model scales and reasoning benchmarks show that \textsc{MicroMouse} consistently outperforms Full-CoT SFT and existing selective-SFT baselines, yielding average relative gains of \textbf{5.5\%} in greedy accuracy and \textbf{2.6\%} in Pass@1 relative to Full-CoT SFT. Meanwhile, \textsc{MicroMouse} improves reasoning efficiency by reducing average output length by \textbf{8.8\%}, while identifying harmful supervision in only \textbf{1.7} seconds per long CoT on average.
\end{abstract}

\section{Introduction}
Recent large reasoning models (LRMs) have demonstrated strong performance on complex reasoning tasks by generating extended chains of thought (CoTs) and leveraging increased inference-time computation \citep{DBLP:journals/corr/abs-2412-16720,Guo_2025}. 
CoTs enable models to solve complex problems through verbalized intermediate reasoning steps, substantially improving performance across reasoning-intensive tasks \citep{NEURIPS2022_9d560961}. 
Long CoTs have increasingly become an important source of supervision for supervised fine-tuning (SFT), where student models learn reasoning behaviors from high-quality reasoning trajectories \citep{ye2025limo}.

However, prevailing full-CoT SFT optimizes all tokens in a reasoning trajectory uniformly, regardless of their actual learning value. This treatment can be suboptimal for long-form CoTs, as even successful trajectories may contain harmful supervision such as erroneous exploration and redundant wandering. Indiscriminately learning from these steps can reinforce undesirable reasoning patterns, thereby degrading both reasoning performance and inference efficiency \citep{chen2025do, cui-etal-2025-stepwise, li2026making}. This motivates an automated framework that identifies such harmful supervision and selectively masks its contribution to the SFT objective.

\begin{figure}[t]
\centering
\counterwithout{figure}{section}

\includegraphics[width=1.0\linewidth]{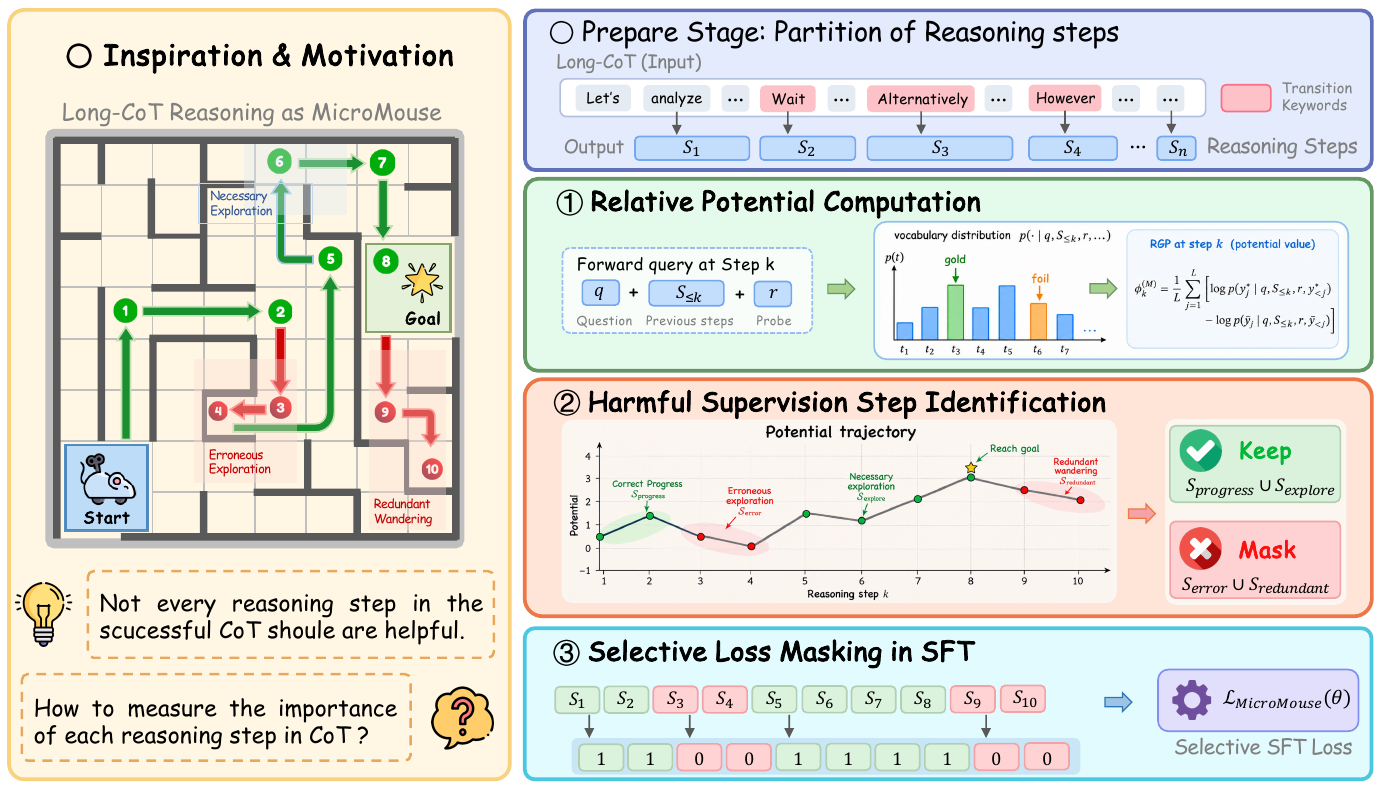}  
\caption{\textbf{Overview of MicroMouse}, a framework consisting of three stages.}
\label{fig:method}
\end{figure}

As shown in Figure~\ref{fig:method} (left), our motivation is inspired by the \emph{MicroMouse} maze-solving problem \citep{520440}. During the search phase, a MicroMouse explores an unknown maze through a sequence of movements, where erroneous forward and backtracking decisions (i.e., passing the goal and returning) may occur. During the dash phase, it instead follows the computed fastest route to the destination. We draw an analogy between this search trajectory and a long CoT: reasoning steps may likewise contain erroneous exploration or redundant wandering after the target answer has been reached. This inspires a new perspective: maintaining a goal-relative measure throughout reasoning, analogous to a Micromouse's spatial coordinates, to quantify how close each reasoning step is to the target answer and guide SFT to mask erroneous exploration and redundant wandering.

Building on this intuition, we propose \textsc{MicroMouse}, an automated selective-SFT framework for masking harmful supervision in long CoTs. Given a long CoT, we first partition it into a sequence of reasoning steps using naturally occurring transition keywords. We then introduce the \emph{Relative Potential} (REP) to quantify how close each reasoning state is to the target answer. Perplexity provides a simple and intuitive measure of how strongly a reasoning prefix supports the gold answer, but the gold answer's perplexity alone captures only absolute confidence and ignores how competing answers change. REP instead contrasts the gold answer with a matched foil through the difference between their log perplexities, equivalently the log-ratio of their perplexities. This relative formulation cancels likelihood variations shared by the matched pair and more directly tracks changes in the model's preference for the target answer. By tracking the evolution of REP along the trajectory, we systematically categorize reasoning steps into Correct Progress, Necessary Exploration, Erroneous Exploration, and Redundant Wandering. During SFT, tokens belonging to erroneous or redundant steps are masked from the loss while remaining visible in the context.


Extensive experiments show that \textsc{MicroMouse} consistently outperforms Full-CoT SFT across student models of different scales and a broad range of reasoning benchmarks. Averaged over Qwen3-4B, Qwen3-8B, and Qwen3-14B, \textsc{MicroMouse} achieves relative improvements of \textbf{5.5\%} in accuracy under greedy decoding and \textbf{2.6\%} in Pass@1 under temperature sampling, while also clearly surpassing existing selective-SFT baselines. In addition, \textsc{MicroMouse} reduces the average output length by \textbf{8.8\%}. The harmful supervision identification is also efficient, requiring only \textbf{1.7} seconds on average for a long CoT containing 22 reasoning steps.


\section{Methodology}
\label{section2}
\subsection{Preliminary}
\paragraph{Problem Definition.}
We focus on reasoning tasks with objectively verifiable target answers, including mathematical reasoning and multiple-choice reasoning. A long-form CoT \(\mathcal{T}=\{S_1, S_2\ldots,S_n\}\) typically consists of multiple reasoning steps serving different roles, such as problem decomposition, intermediate derivation, exploration, self-correction, and verification. Even when the CoT eventually reaches the correct answer, it may still contain harmful supervision, such as persistent erroneous exploration and redundant wandering after the solution has effectively been reached. Indiscriminately optimizing all such steps may reinforce undesirable reasoning behaviors and impair both reasoning quality and efficiency. Our goal is to identify reasoning steps that should not contribute to supervision and selectively exclude their token-level losses during SFT.

\paragraph{Relative Potential.}
Relative Potential (REP) is designed to measure how close the current reasoning state is to the gold answer along a long CoT. Prior work \citep{xiao2025limopro, cui-etal-2025-stepwise} often uses the perplexity of the gold answer to estimate how strongly a reasoning prefix supports the target answer. However, gold-answer perplexity alone only reflects the absolute likelihood of the target and does not capture how the probability of other answers changes. In particular, the gold answer may become more likely while a nearby incorrect answer increases even more, in which case perplexity would suggest improved progress despite a deterioration in the model's relative preference for the gold answer. Motivated by this limitation, we instead adopt a contrastive view and compare the conditional log-likelihood of the gold answer with that of a foil answer \citep{warstadt-etal-2020-blimp-benchmark}. The foil is constructed as a minimal contrast to the gold answer: it preserves the answer format and token length while introducing a minimal rule-based perturbation \citep{gardner-etal-2020-evaluating} that makes the answer incorrect (see details in Appendix~\ref{app:foil-and-noise}), e.g., \(18\!\rightarrow\!19\), \(A\!\rightarrow\!B\), or \(\frac{4}{5}\!\rightarrow\!\frac{3}{5}\). Given a scoring model \(M\), we define the \emph{Relative Potential} at reasoning step \(k\) as
\begin{equation}
\phi_k^{(M)}
=
\frac{1}{L}
\sum_{j=1}^{L}
\left[
\log p_M\!\left(
y_j^{*}\mid q,S_{\leq k},r,y_{<j}^{*}
\right)
-
\log p_M\!\left(
\tilde{y}_j\mid q,S_{\leq k},r,\tilde{y}_{<j}
\right)
\right]
=
-\log
\frac{\mathrm{PPL}^{\mathrm{gold}}_k}
{\mathrm{PPL}^{\mathrm{foil}}_k},
\label{eq:rep}
\end{equation}
where \(q\) denotes the input question, \(r\) is an answer probe, and \(p_M(\cdot)\) denotes the conditional probability assigned by the scoring model \(M\), and \(L\) is the shared token length of the gold answer \(y^{*}\) and its foil \(\tilde{y}\). Thus, \(\phi_k\) represents the normalized conditional log-likelihood advantage of the gold answer over its foil, or equivalently, their log perplexity ratio. A larger REP indicates stronger relative support for the gold answer. Compared with gold-answer PPL alone, REP reduces shared likelihood variations, yielding a more robust estimate of changes in the reasoning state's support for the gold answer. We provide a formal analysis of these advantages of REP in Appendix~\ref{app:rep_analysis}.

\subsection{Overview}
Given a long-form reasoning trajectory, we first partition it into reasoning steps at naturally occurring transition keywords, yielding
\(\mathcal{T}=\{S_1,S_2,\ldots,S_n\}\).
\textsc{MicroMouse} then identifies harmful reasoning steps for supervision through three stages, as shown in Figure~\ref{fig:method}.

\textbf{Stage 1: Relative Potential Computation.}
We compute the Relative Potential at each reasoning step to estimate the progress of the current reasoning state toward the target answer.

\textbf{Stage 2: Harmful Supervision Step Identification.}
We analyze the changes in REP across reasoning steps to specifically identify erroneous exploration and redundant wandering.

\textbf{Stage 3: Selective Loss Masking in SFT.}
Based on the identified reasoning steps, we construct a selective supervision mask and apply it to the loss optimization of downstream SFT.

\subsection{Stage 1: Relative Potential Computation}

Given a segmented reasoning CoT
\(\mathcal{T}=\{S_1,S_2,\ldots,S_n\}\),
we compute the Relative Potential for each reasoning step sequentially.
For the \(k\)-th reasoning step, we first construct the input query by concatenating the original question \(q\), the reasoning prefix \(S_{\leq k}\), and a fixed answer probe \(r\), instantiated as
\texttt{</think> So, the final answer is \textbackslash boxed\{}. The resulting input can be written as $[q;\,S_{\leq k};\,r].$.
We feed it into the scoring model \(M\) and compute the REP \(\phi_k^{(M)}\) from the conditional log-likelihood difference between the gold answer \(y^{*}\) and its foil \(\tilde{y}\), following Eq.~\ref{eq:rep}.
Repeating this procedure for \(k=1,\ldots,n\) yields the complete potential trajectory:
\begin{equation}
\Phi^{(M)}
=
\{\phi_1^{(M)},\phi_2^{(M)},\ldots,\phi_n^{(M)}\},
\end{equation}
which characterizes the evolution of the reasoning state's proximity to the target answer.

\subsection{Stage 2: Harmful Supervision Step Identification}

Given the REP trajectory
\(\Phi^{(M)}=\{\phi_1^{(M)},\phi_2^{(M)},\ldots,\phi_n^{(M)}\}\),
we use the potential change
\(\Delta\phi_k^{(M)}=\phi_k^{(M)}-\phi_{k-1}^{(M)}\)
to quantify the effect of reasoning step \(S_k\), and define
\(k^{\max}=\arg\max_k \phi_k^{(M)}\)
as the step reaching the maximum REP.
Let \(\sigma\) denote the REP noise scale estimated from the lag-1 autocovariance of its increments (see details in Appendix~\ref{app:foil-and-noise}).We categorize all reasoning steps into four categories:

\textbf{(a) Correct Progress:}
\begin{equation}
    \mathcal{S}_{\mathrm{progress}}
=
\left\{
S_k \mid
k\leq k^{\max},
\;
\Delta\phi_k^{(M)}\geq-\sigma
\right\}.
\end{equation}
These steps either increase the REP or introduce only minor fluctuations within the noise scale, and are retained for supervision.

\textbf{(b) Necessary Exploration:}
\begin{equation}
\mathcal{S}_{\mathrm{explore}}
=
\left\{
S_k \mid
k\leq k^{\max},
\;
\Delta\phi_k^{(M)}<-\sigma,
\;
\Delta\phi_{k-1}^{(M)}\geq-\sigma,
\;
\Delta\phi_{k+1}^{(M)}\geq-\sigma
\right\}.
\end{equation}
These isolated potential decreases may reflect necessary local deviations toward the target answer, so we conservatively retain them for supervision.

\textbf{(c) Erroneous Exploration:}
\begin{equation}
\mathcal{S}_{\mathrm{error}}
=
\left\{
S_k \mid
k\leq k^{\max},
\;
\Delta\phi_k^{(M)}<-\sigma,
\;
\left(
\Delta\phi_{k-1}^{(M)}<-\sigma
\;\lor\;
\Delta\phi_{k+1}^{(M)}<-\sigma
\right)
\right\}.
\end{equation}
These steps belong to consecutive potential decreases, indicating persistent movement away from the target answer, and are masked.

\textbf{(d) Redundant Wandering:}
\begin{equation}
\mathcal{S}_{\mathrm{redundant}}
=
\left\{
S_k \mid
k>k^{\max}
\right\}.
\end{equation}
These steps occur after the highest-potential reasoning step and may correspond to redundant reasoning that fails to terminate in time, and are therefore masked.

Accordingly, \textsc{MicroMouse} keeps
\(\mathcal{S}_{\mathrm{progress}}\cup\mathcal{S}_{\mathrm{explore}}\)
while masking
\(\mathcal{S}_{\mathrm{error}}\cup\mathcal{S}_{\mathrm{redundant}}\).

\subsection{Stage 3: Selective Loss Masking in SFT}

Directly removing harmful reasoning steps from the training trajectory may break the continuity of the reasoning CoT and disrupt the contextual dependencies required by subsequent reasoning. We therefore preserve the complete reasoning CoT and selectively mask the corresponding token losses during SFT. Given an input token sequence
\(T =\{t_1,t_2,\ldots,t_l\}\),
the \textsc{MicroMouse} objective is formulated as
\begin{equation}
\mathcal{L}_{\mathrm{MicroMouse(\theta)}}
=
-
\frac{1}{\sum_{i=1}^{l}\mathbb{I}(t_i)}
\sum_{i=1}^{l}
\mathbb{I}(t_i)
\log
P_{\theta}
\left(
t_i \mid q,t_{<i};\theta
\right),
\label{eq:MicroMouse_loss}
\end{equation}
where \(\mathbb{I}(t_i)=0\) if \(t_i\) belongs to a reasoning step in \(\mathcal{S}_{\mathrm{error}}\cup\mathcal{S}_{\mathrm{redundant}}\), and \(\mathbb{I}(t_i)=1\) otherwise.




\section{Experiments}
\label{section3}
\subsection{Experimental Setup}
\paragraph{Datasets.}
We use the official LIMO training set \citep{ye2025limo}, which contains 817 carefully curated mathematical problems paired with long-form reasoning CoTs, as the supervision data for all SFT experiments. Following a similar evaluation protocol to \citet{ye2025limo}, we evaluate the resulting models on a diverse suite of reasoning benchmarks covering both in-domain and out-of-domain generalization. The in-domain benchmarks include MATH500 \citep{DBLP:journals/corr/abs-2103-03874}, AMC23, and AIME24, which closely match the mathematical reasoning distribution of the training data. We further include HMMT26 as a recent competition-mathematics benchmark to assess generalization to newly released and unseen problems. For out-of-domain evaluation, we consider GPQA-Diamond \citep{rein2023gpqagraduatelevelgoogleproofqa}, Minerva \citep{lewkowycz2022solving}, and OlympiadBench \citep{he-etal-2024-olympiadbench}, which cover challenging scientific, quantitative, and Olympiad-level reasoning problems. 

\paragraph{Metrics.}
We evaluate reasoning performance under greedy decoding and temperature sampling. Greedy decoding uses exact-match accuracy (\textbf{Acc}), while temperature sampling reports \textbf{Pass@1} and \textbf{Pass@6}, measuring average correctness and the probability of obtaining at least one correct solution within six samples, respectively.

\paragraph{Implementation Details.}
During training, we use DeepSeek-R1-Distill-Qwen-1.5B as the default scoring model \(M\) to compute the Relative Potential and derive the selective supervision masks. We conduct SFT on three dense Qwen3 models, namely Qwen3-4B-Base, Qwen3-8B-Base, and Qwen3-14B-Base. All models are trained for 9 epochs on the 817 long-CoT trajectories from LIMO with a learning rate of \(1\times10^{-5}\). For a fair comparison, Full CoT SFT, \textsc{MicroMouse}, and other existing baselines use exactly the same training data and hyperparameters. During evaluation, we evaluate all models under a zero-shot CoT setting with a maximum generation length of 32,768 tokens. Greedy decoding uses \(T=0\), while temperature sampling uses \(T=0.6\). For MATH500, GPQA-Diamond, Minerva, and OlympiadBench, we sample 6 responses per problem. For the smaller AMC23, AIME24, and HMMT26 benchmarks, we increase the number of sampled responses to 16 per problem to obtain a more stable estimate of average performance. All training and evaluation experiments are conducted on 8 NVIDIA H100 GPUs.

\subsection{Main Results}
Table~\ref{tab:main} presents a comparison across seven reasoning benchmarks, three model scales, and two decoding settings. Compared with the corresponding base models before SFT, both fine-tuning strategies substantially improve reasoning performance, while \textsc{MicroMouse} delivers more consistent gains over Full CoT SFT across evaluation settings. Specifically, averaged across the three model scales, \textsc{MicroMouse} achieves a relative improvement of \textbf{5.5\%} in accuracy under greedy decoding, and relative improvements of \textbf{2.6\%} and \textbf{0.9\%} in Pass@1 and Pass@6 under temperature sampling. The gains are particularly pronounced for Qwen3-8B, where \textsc{MicroMouse} improves over Full CoT SFT by \textbf{9.9\%} under greedy decoding and \textbf{4.2\%} in Pass@1 under temperature sampling. 

We further observe that the improvement is generally larger under greedy decoding than under temperature sampling. One plausible explanation is that greedy decoding directly reflects changes to the model's highest-probability reasoning trajectory, whereas stochastic sampling explores a broader set of candidate trajectories and consequently reduces the sensitivity of aggregate metrics to improvements concentrated around the dominant decoding path. Nevertheless, \textsc{MicroMouse} retains consistent gains under temperature sampling,  indicating that its benefits are not restricted to a single deterministic decoding trajectory and remain robust under more diverse generation settings.

\begin{table}[t]
\centering
\footnotesize
\setlength{\tabcolsep}{3.0pt}
\renewcommand{\arraystretch}{1.08}

\caption{
\textbf{Main results across model scales and reasoning benchmarks.}
We compare Base, Full CoT SFT, and \textsc{MicroMouse} on four in-domain
and three out-of-domain benchmarks.
Overall reports accuracy under greedy decoding and Pass@1/Pass@6
under temperature sampling. ``--'' indicates that Pass@6 is not applicable to greedy decoding.
The best results are highlighted in \textbf{bold}.
}
\label{tab:main}

\begin{adjustbox}{max width=\linewidth}
\begin{tabular}{@{}ll cccc ccc cc@{}}
\toprule
&& \multicolumn{4}{c}{\textbf{In-Domain}}
& \multicolumn{3}{c}{\textbf{Out-of-Domain}}
& \multicolumn{2}{c}{\textbf{Overall}} \\
\cmidrule(lr){3-6}
\cmidrule(lr){7-9}
\cmidrule(lr){10-11}

\textbf{Model} & \textbf{Method}
& \textbf{MATH500}
& \textbf{AMC23}
& \textbf{AIME24}
& \textbf{HMMT26}
& \textbf{GPQA}
& \textbf{Minerva}
& \textbf{Olympiad}
& \textbf{Acc./P@1}
& \textbf{P@6} \\
\midrule

\multicolumn{11}{@{}l}{\textit{Greedy Decoding ($T=0$)}} \\
\midrule

\multirow{3}{*}{Qwen3-4B}
& Base
& 62.6 & 47.5 & 16.7 & 3.0
& 17.2 & 19.9 & 32.6 
& 28.5 & -- \\

& Full CoT SFT
& 70.8 & 47.5 & 13.3 & 12.1
& 35.4 & \textbf{21.3} & 38.8
& 34.2 & -- \\

& \textsc{MicroMouse}
& \textbf{71.6} & \textbf{55.0} & \textbf{13.3} & \textbf{12.1}
& \textbf{36.9} & 21.0 & \textbf{39.3}
& \textbf{35.6}\gain{4.2} & -- \\

\cmidrule(lr){1-11}

\multirow{3}{*}{Qwen3-8B}
& Base
& 76.2 & 62.5 & 13.3 & 9.1
& 41.4 & 29.4 & 40.3
& 38.9 & -- \\

& Full CoT SFT
& 76.0 & 57.5 & 30.0 & 9.1
& 42.9 & 30.1 & 41.6 
& 41.0 & -- \\

& \textsc{MicroMouse}
& \textbf{80.4} & \textbf{70.0} & \textbf{30.0} & \textbf{12.1}
& \textbf{44.9} & \textbf{30.9} & \textbf{47.4} 
& \textbf{45.1}\gain{9.9} & -- \\

\cmidrule(lr){1-11}

\multirow{3}{*}{Qwen3-14B}
& Base
& 81.0 & 65.0 & 13.3 & 12.1
& 42.9 & 32.0 & 44.9
& 41.6 & -- \\

& Full CoT SFT
& \textbf{91.4} & 85.0 & \textbf{50.0} & 24.2
& 55.6 & \textbf{36.0} & \textbf{60.4}
& 57.5 & -- \\

& \textsc{MicroMouse}
& 90.8 & \textbf{87.5} & 43.3 & \textbf{36.4}
& \textbf{58.6} & \textbf{36.0} & 59.9
& \textbf{58.9}\gain{2.5} & -- \\





\midrule
\multicolumn{11}{@{}l}{\textit{Temperature Sampling ($T=0.6$)}} \\
\midrule

\multirow{3}{*}{Qwen3-4B}
& Base
& 35.8 & 26.6 & 7.2 & 3.2
& 7.2 & 12.1 & 20.0
& 16.0 & 40.5 \\

& Full CoT SFT
& \textbf{81.0} & \textbf{65.3} & 20.0 & 11.9
& 41.1 & \textbf{30.9} & \textbf{47.3}
& 42.5 & 62.2 \\

& \textsc{MicroMouse}
& 80.5 & 65.0 & \textbf{23.8} & \textbf{15.2}
& \textbf{42.6} & 28.9 & 46.4
& \textbf{43.2}\gain{1.6}
& \textbf{62.4}\gain{0.3} \\

\cmidrule(lr){1-11}

\multirow{3}{*}{Qwen3-8B}
& Base
& 72.9 & 52.5 & 11.9 & 5.1
& 40.7 & 26.0 & 37.3
& 35.2 & 57.2 \\

& Full CoT SFT
& 84.4 & 71.4 & 30.6 & 19.1
& 48.7 & 34.0 & 52.3
& 48.6 & 68.9 \\

& \textsc{MicroMouse}
& \textbf{86.5} & \textbf{73.4} & \textbf{32.9} & \textbf{21.8}
& \textbf{51.5} & \textbf{35.8} & \textbf{54.5}
& \textbf{50.9}\gain{4.7}
& \textbf{70.3}\gain{2.0} \\

\cmidrule(lr){1-11}

\multirow{3}{*}{Qwen3-14B}
& Base
& 73.8 & 53.0 & 12.9 & 8.9
& 42.8 & 29.6 & 38.8
& 37.1 & 59.1 \\

& Full CoT SFT
& \textbf{91.5} & \textbf{85.2} & 46.9 & 29.7
& 57.3 & 35.2 & \textbf{61.1}
& 58.1 & 73.5 \\

& \textsc{MicroMouse}
& 91.3 & 85.0 & \textbf{50.8} & \textbf{31.1}
& \textbf{59.9} & \textbf{35.3} & 59.6
& \textbf{59.0}\gain{1.5}
& \textbf{73.8}\gain{0.4} \\





\bottomrule
\end{tabular}
\end{adjustbox}
\end{table}

\paragraph{Performance across Problem Difficulty.}
To investigate whether \textsc{MicroMouse} provides different benefits across problem difficulty, we further break down the MATH500 results on Qwen3-8B according to the five difficulty levels. Following the easy-to-hard split adopted by \citet{sun2024easytohardgeneralizationscalablealignment}, we group Levels 1--3 as easy problems and Levels 4--5 as hard problems, as shown in Table~\ref{tab:math500_difficulty}. Overall, \textsc{MicroMouse} improves over Full CoT SFT from 84.4\% to 86.5\% in Pass@1 and from 95.4\% to 96.0\% in Pass@6, corresponding to relative improvements of approximately 2.5\% and 0.6\%. More importantly, the benefit becomes more pronounced as problem difficulty increases. \textsc{MicroMouse} improves Pass@1 over Full CoT SFT by only 0.4\% on easy problems, but by 4.8\% on hard problems, indicating substantially larger gains as problem difficulty increases. A similar tendency is observed for Pass@6, although the margin is smaller due to the near-saturated performance. This pattern is consistent with our intuition that harder problems generally involve more complex and exploratory reasoning, increasing the likelihood of persistent erroneous exploration and redundant reasoning. By selectively suppressing supervision from such reasoning steps, \textsc{MicroMouse} provides a larger advantage on problems that require more challenging reasoning trajectories.

\begin{table}[t]
\centering
\small
\setlength{\tabcolsep}{3.0pt}
\renewcommand{\arraystretch}{1.08}
\caption{
\textbf{Qwen3-8B temperature sampling performance across MATH500 difficulty levels.}
}
\label{tab:math500_difficulty}

\begin{tabular}{@{}ll cccc ccc c@{}}
\toprule
\multirow{2}{*}{\textbf{Setting}}
& \multirow{2}{*}{\textbf{Metric}}
& \multicolumn{4}{c}{\textbf{Easy}}
& \multicolumn{3}{c}{\textbf{Hard}}
& \multirow{2}{*}{\textbf{Total Avg.}} \\
\cmidrule(lr){3-6}
\cmidrule(lr){7-9}
&
& \textbf{Level 1} & \textbf{Level 2} & \textbf{Level 3} & \textbf{Avg.}
& \textbf{Level 4} & \textbf{Level 5} & \textbf{Avg.}
& \\
\midrule

\multirow{2}{*}{Full CoT SFT}
& Pass@1
& 89.1 & \textbf{93.9} & 91.0 & 91.8
& 82.6 & 73.1 & 77.7
& 84.4 \\
& Pass@6
& 95.3 & 98.9 & \textbf{99.0} & 98.3
& 95.3 & 90.3 & 92.7
& 95.4 \\

\midrule

\multirow{2}{*}{\textsc{MicroMouse}}
& Pass@1
& \textbf{92.6} & 92.0 & \textbf{92.1} & \textbf{92.2}\gain{0.4}
& \textbf{85.8} & \textbf{77.2} & \textbf{81.4}\gain{4.8}
& \textbf{86.5}\gain{2.5} \\
& Pass@6
& \textbf{100.0} & \textbf{98.9} & 98.1 & \textbf{98.7}\gain{0.4}
& \textbf{96.1} & \textbf{91.0} & \textbf{93.5}\gain{0.9}
& \textbf{96.0}\gain{0.6} \\

\bottomrule
\end{tabular}
\end{table}


\subsection{Comparison with Exiting Methods}

\begin{figure}[t]
\centering
\counterwithout{figure}{section}

\includegraphics[width=1.0\linewidth]{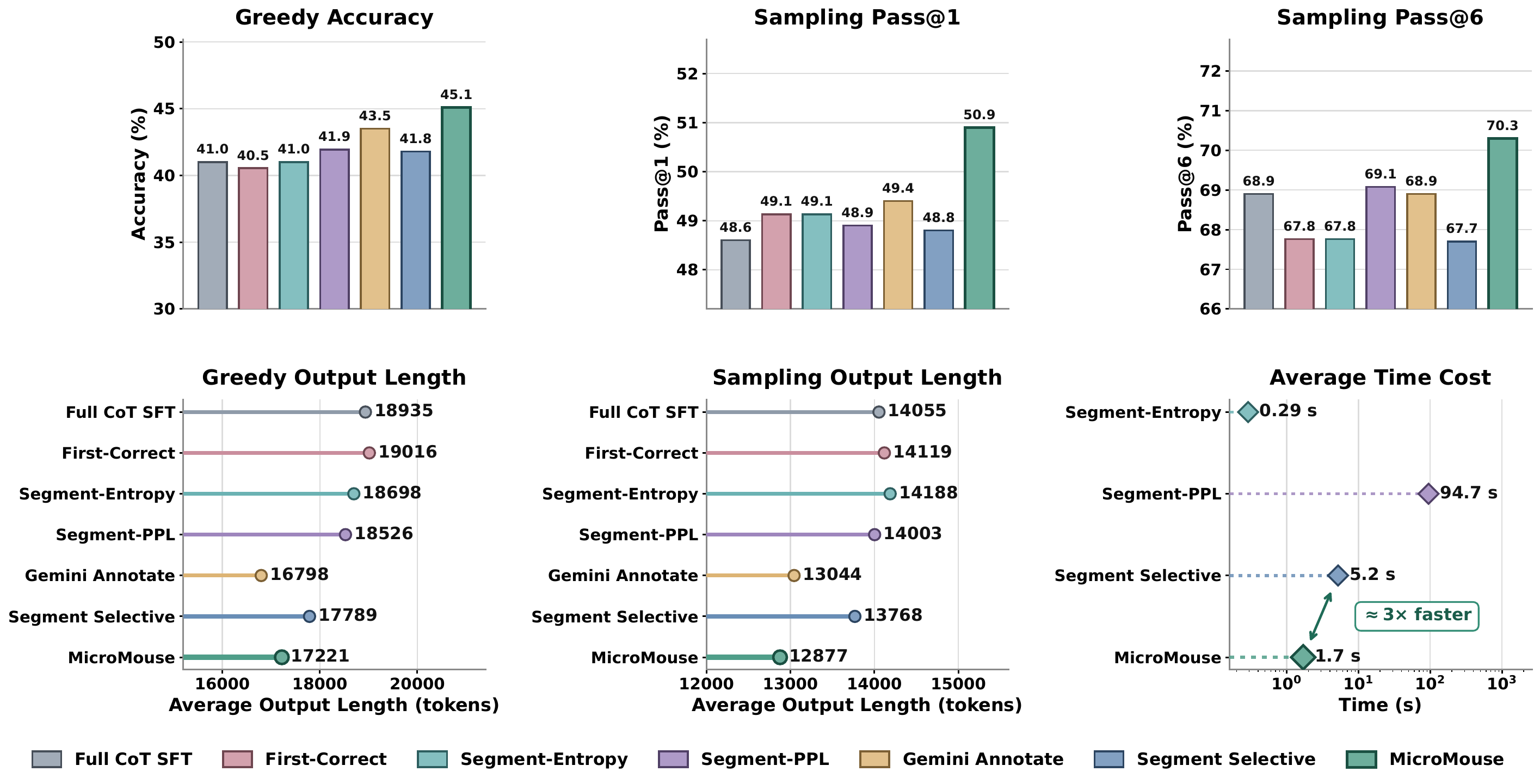}  
\caption{
\textbf{Comparison with other selective-SFT baselines on Qwen3-8B.}
The top row reports reasoning performance under greedy decoding and temperature sampling, while the bottom row compares output length and the computational cost of masking harmful supervision.
}
\label{fig:comparison}
\end{figure}


In this section, we compare \textsc{MicroMouse} on Qwen3-8B against five selective-SFT baselines: (1) \textbf{First-Correct} \citep{chen2025do}, which retains the segment where the correct answer first appears together with all preceding segments; (2) \textbf{Segment-Entropy} \citep{li2026making}, which prioritizes segments with higher aggregated token-level entropy, indicating greater information contribution; (3) \textbf{Segment-PPL} \citep{cui-etal-2025-stepwise}, which estimates segment importance by measuring the increase in CoT perplexity after removing a reasoning segment; (4) \textbf{Gemini Annotate}, which directly uses Gemini-3.6-Flash to annotate the importance of each reasoning step, with the annotation prompt provided in the Appendix~\ref{app:annotation_prompt}; and (5) \textbf{Segment Selective} \citep{wang2026segmentlevel}, which identifies important segments by aggregating token-level Integrated Gradients (IG) signals. For Segment-Entropy, Segment-PPL, Segment Selective, and \textsc{MicroMouse}, we use the same scoring model and match the supervision budget to ensure a fair comparison.

As shown in the upper row of Figure~\ref{fig:comparison}, we compare \textsc{MicroMouse} with Full CoT SFT and five selective-SFT baselines across seven reasoning benchmarks. Several selective-SFT strategies consistently outperform Full CoT SFT, providing further evidence that long CoTs contain harmful supervision and that masking such signals can yield stable gains in reasoning performance. Among the six selective-SFT methods, \textsc{MicroMouse} achieves the best overall performance under both decoding settings. These results further suggest that measuring each reasoning step by its relative potential enables more accurate identification and masking of harmful supervision in long CoTs. Interestingly, Gemini Annotate achieves the second-best overall performance, indicating that direct reasoning-step annotation with a strong LLM is also effective for selective supervision.


\paragraph{Reasoning Efficiency.}
The bottom row of Figure~\ref{fig:comparison} further compares the average output length of Full CoT SFT and the six selective-SFT strategies under both greedy decoding and temperature sampling. First-Correct, Segment-Entropy, and Segment-PPL produce output lengths comparable to Full CoT SFT, whereas Gemini Annotate, Segment Selective, and \textsc{MicroMouse} yield noticeably shorter generations. In particular, \textsc{MicroMouse} reduces the average output length by 9.1\% under greedy decoding and 8.4\% under temperature sampling relative to Full CoT SFT.  Notably, \textsc{MicroMouse-SFT} improves reasoning performance while producing shorter generations. This joint improvement suggests that masking harmful supervision during SFT encourages the student model to learn more effective and concise reasoning behaviors, reducing erroneous exploration and redundant wandering while improving reasoning performance. Although Gemini Annotate also reduces output length, \textsc{MicroMouse} achieves stronger performance, resulting in a better trade-off.

\paragraph{Computational Overhead.}
The bottom-right panel of Figure~\ref{fig:comparison} compares the average time cost of the annotation-free methods on all 817 LIMO trajectories, each containing about 22 reasoning steps. Segment-Entropy is the fastest because it requires only one forward pass to compute token-level entropy, whereas Segment-PPL repeatedly removes reasoning steps and recomputes perplexity, resulting in a much higher cost of 94.7 seconds per trajectory. Segment Selective takes 5.2 seconds due to token-level IG computation and aggregation. In contrast, \textsc{MicroMouse} only evaluates the conditional log-likelihoods of the gold answer and its matched foil at each step, requiring 1.7 seconds per trajectory and running about \(3\times\) faster than Segment Selective. Although Segment-Entropy is cheaper, \textsc{MicroMouse} yields substantially stronger downstream performance, achieving a better balance between supervision quality and computational cost for large scale long CoT SFT.
\subsection{Ablation Study}

\begin{wraptable}{r}{0.50\textwidth}
\vspace{-1.0\baselineskip}
\centering
\footnotesize
\setlength{\tabcolsep}{2.6pt}
\renewcommand{\arraystretch}{1.08}

\caption{
\textbf{Ablation studies on Qwen3-8B.}
}
\label{tab:ablation_study}

\resizebox{\linewidth}{!}{
\begin{tabular}{@{}l c cc ccc@{}}
\toprule
& & \multicolumn{2}{c}{\textbf{Greedy}}
& \multicolumn{3}{c}{\textbf{Sampling}} \\
\cmidrule(lr){3-4}
\cmidrule(lr){5-7}

\textbf{Variant}
& \textbf{Keep (\%)}
& \textbf{Acc.}
& \textbf{Len. (k)}
& \textbf{P@1}
& \textbf{P@6}
& \textbf{Len. (k)} \\
\midrule

Full CoT SFT
& 100.0
& 41.0
& 18.94
& 48.6
& 68.9
& 14.06 \\

Random Mask
& 80.9
& 41.7
& 17.89
& 49.1
& 68.2
& 14.01 \\

\rowcolor{gray!10}
\textsc{MicroMouse}
& 80.9
& \textbf{45.1}
& 17.22
& \textbf{50.9}
& \textbf{70.3}
& 12.88 \\

\midrule

\quad w/o Mask $\mathcal{S}_{\mathrm{error}}$
& 83.5
& 44.4
& 16.67
& 50.7
& 70.2
& 12.84 \\

\quad w/o Mask $\mathcal{S}_{\mathrm{redundant}}$
& 97.1
& 38.9
& 19.34
& 49.0
& 69.0
& 14.11 \\

\quad w/ Mask $\mathcal{S}_{\mathrm{explore}}$
& 74.1
& 41.8
& 18.06
& 48.9
& 68.2
& 13.81 \\

\midrule

\quad w/ Qwen3-4B
& 85.5
& 41.9
& 18.27
& 49.0
& 70.2
& 14.18 \\

\bottomrule
\end{tabular}
}
\vspace{-1.0\baselineskip}
\end{wraptable}

We investigate the contributions of individual \textsc{MicroMouse} components and the effect of replacing the scoring model, with results summarized in Table~\ref{tab:ablation_study}. For component ablations, we consider four variants: (1) \textit{Random Mask}, which randomly masks reasoning segments until matching the supervision budget of \textsc{MicroMouse}; (2) \textit{w/o Mask $\mathcal{S}_{\mathrm{error}}$}, which does not mask reasoning steps identified as erroneous exploration; (3) \textit{w/o Mask $\mathcal{S}_{\mathrm{redundant}}$}, which does not mask reasoning steps identified as redundant wandering; and (4) \textit{w/ Mask $\mathcal{S}_{\mathrm{explore}}$}, which additionally masks reasoning segments exhibiting a single-step decrease in REP. We further replace the default DeepSeek-R1-Distill-Qwen-1.5B with Qwen3-4B-Instruct to assess the sensitivity of \textsc{MicroMouse} to the choice of scoring model.

The component ablations reveal that the gains of \textsc{MicroMouse} arise primarily from which reasoning steps are excluded from supervision, rather than from simply reducing the amount of training signal. Random Mask, despite using the same supervision budget as \textsc{MicroMouse}, yields only marginal changes over Full CoT SFT, while remaining substantially below \textsc{MicroMouse} across all performance metrics. Removing the masking of either erroneous exploration or redundant wandering also degrades performance, supporting the necessity of suppressing these two types of harmful supervision. In particular, failing to mask redundant wandering results in relative performance drops of 13.7\% in greedy accuracy, 3.4\% in Pass@1, and 1.9\% in Pass@6 under temperature sampling. Conversely, additionally masking reasoning steps with only a single-step decrease in REP causes a marked performance degradation. This result supports our design choice of preserving isolated potential decreases, which may correspond to necessary intermediate exploration rather than genuinely harmful supervision signals. Finally, replacing the default scoring model with Qwen3-4B-Instruct still yields consistent improvements over Full CoT SFT, including a 1.9\% relative gain in Pass@6. This result indicates that the effectiveness of \textsc{MicroMouse} is not restricted to the default scorer and remains stable when using another alternative scoring model.

\subsection{Case Study}

\begin{figure}[t]
\centering
\counterwithout{figure}{section}

\includegraphics[width=1.0\linewidth]{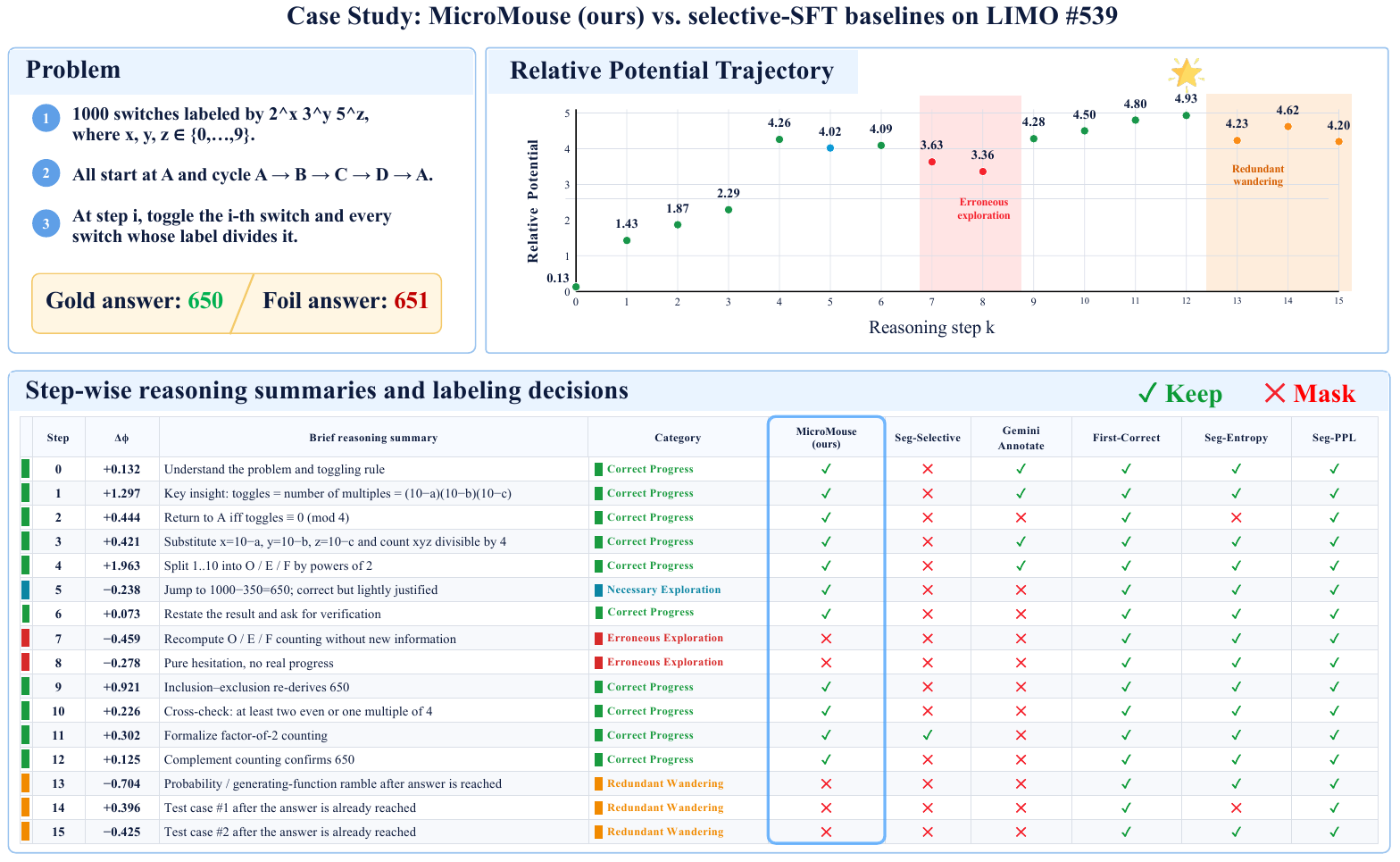}  
\caption{
\textbf{Case study of step-wise supervision analysis on LIMO \#539.}
The figure visualizes the Relative Potential trajectory and compares the supervision decisions of \textsc{MicroMouse} with selective-SFT baselines across all reasoning steps.
}
\label{fig:case_study}
\end{figure}

Figure~\ref{fig:case_study} visualizes the Relative Potential trajectory of \textsc{MicroMouse} on LIMO sample \#539 and compares its step-wise supervision decisions with those of other selective-SFT baselines. The original reasoning trajectory is provided in Appendix~\ref{app:case_study_limo539}. Several observations emerge from this example.
\textbf{First, an isolated decrease in Relative Potential does not necessarily indicate harmful supervision.} At Step~5, the model reaches the correct answer through a logical jump, causing a temporary decrease in Relative Potential. Although the argument at this step is insufficiently justified, it still advances toward the target answer, and the subsequent step begins to further verify the result. 
\textbf{Second, consecutive decreases in Relative Potential correspond more clearly to erroneous exploration.} At Steps~7--8, the reasoning repeatedly recomputes an already established counting argument and then transitions into hesitation without introducing new progress. 
\textbf{Third, reasoning steps after the Relative Potential reaches its peak exhibit clear redundant wandering.} By Step~12, the model has repeatedly cross-validated the derivation and reached the target answer with high confidence. Nevertheless, Steps~13--15 continue to revisit alternative formulations and perform additional sanity checks, despite no longer contributing substantive progress toward the solution.

Taken together, this example provides empirical support for \textsc{MicroMouse}'s identification of erroneous exploration and redundant wandering based on the evolution of REP. It also illustrates why isolated potential drops should be treated conservatively, as they may correspond to necessary exploration rather than genuinely harmful supervision. In contrast, other selective-SFT baselines are either overly aggressive or permissive, leading to less accurate supervision selection on this example.

\section{Related Work}
\paragraph{Reasoning Efficiency of Large Reasoning Models.}
Large reasoning models (LRMs) achieve strong performance through extended chains of thought (CoTs), but long-form reasoning often incurs substantial computational overhead and redundant reasoning \citep{sui2025stop,wang2026harnessingreasoningeconomysurvey}. Existing approaches improve reasoning efficiency either at inference time or through training. Prompt- and budget-based methods explicitly constrain test-time computation \citep{han-etal-2025-token,Renze_2024}, while latent-reasoning methods reduce explicit verbalized computation by performing intermediate reasoning in continuous representations \citep{hao2025training}. Training-based methods instead refine or compress CoT supervision, for example by identifying critical reasoning steps using perplexity-based criteria \citep{cui-etal-2025-stepwise}. In contrast, \textsc{MicroMouse} targets the quality rather than the length of long-CoT supervision, preserving the complete trajectory as context while excluding harmful reasoning units from loss optimization.

\paragraph{Selective SFT and Importance Measurement for Reasoning CoTs.}
Recent work challenges the conventional SFT objective that optimizes all target tokens uniformly. Rho-1 selectively applies loss to high-utility tokens identified by a reference model \citep{lin2025rho1tokensneed}, Goldfish Loss randomly excludes subsets of target tokens from optimization \citep{hans2024be}, and Token Cleaning removes low-quality token-level supervision during SFT \citep{pang2025token}. At the step level, ATLAS identifies critical steps in agent trajectories and fine-tunes only on selected actions \citep{chen-etal-2025-atlas}, while GLM-5 \citep{glm5team2026glm5vibecodingagentic} masks erroneous trajectory segments while retaining them as context for subsequent correction. For long reasoning CoTs, the central challenge is therefore to determine which parts of a trajectory should contribute to supervision. Existing methods operate primarily at token or segment granularity. Token-level approaches use signals such as logits, entropy, gradients, or contrastive differences to identify informative or error-critical tokens \citep{cheng2025reasoningexplorationentropyperspective,wang2025beyond}. Since individual tokens may not correspond to semantically complete reasoning decisions, segment-level methods instead evaluate coherent reasoning units using perplexity \citep{cui-etal-2025-stepwise}, step entropy \citep{li2026making}, or perturbation-based attribution \citep{xu-etal-2025-procut}, while \citet{wang2026segmentlevel} aggregate token-level Integrated Gradients into segment-level attribution signals. Rather than assigning each segment an isolated importance score, \textsc{MicroMouse} tracks the evolution of Relative Potential along the reasoning trajectory to identify erroneous exploration and redundant wandering, and masks the corresponding harmful supervision during SFT.
\section{Conclusion}
In this work, we propose \textsc{MicroMouse}, an automated selective-SFT framework for masking harmful supervision in long CoTs. \textsc{MicroMouse} maintains a Relative Potential to measure how close each reasoning state is to the target answer, enabling the identification of erroneous exploration and redundant wandering along long reasoning trajectories. Unlike single answer log-perplexity measures, Relative Potential is defined through the conditional log-likelihood difference between the gold answer and a matched foil answer, which reduces shared likelihood fluctuations and provides a more stable signal for tracking reasoning progress. Extensive experiments s show that \textsc{MicroMouse} consistently outperforms Full CoT SFT and achieves stronger overall performance than existing selective-SFT baselines. It also delivers clear advantages in reasoning efficiency and computational overhead. Overall, \textsc{MicroMouse} provides an effective and scalable approach for identifying and masking harmful supervision in large-scale long-CoT SFT.

\bibliography{iclr2027_conference}
\bibliographystyle{iclr2027_conference}

\appendix

\section{Foil Construction and Noise-Scale Estimation}
\label{app:foil-and-noise}

\subsection{Foil construction}

The potential function is read out as a length-normalised log-likelihood ratio
between the gold answer and a \emph{foil}: an answer string that matches the gold
one in token count and surface format but carries a minimally different,
incorrect value. Matching length and format holds everything except
value-correctness fixed, so the ratio measures only whether the reasoning prefix
has pinned down the right value; taking the \emph{nearest} wrong value makes the
foil the hardest available competitor, which keeps the signal sensitive to
reasoning progress instead of saturating early. Answers are numeric: an integer
answer $v$ is perturbed to $v+1$, falling back to $v-1$ whenever $v+1$ would
change the digit count or equal zero, and a fractional answer is perturbed by
incrementing its numerator.

\begin{lstlisting}
import re

def make_foil(ans):
    '''Minimally-perturbed counterfactual answer:
       same digit count, same format, nearest value, but wrong.'''
    a = str(ans).strip()

    m = re.fullmatch(r'-?\d+', a)                      # integer answers
    if m:
        v = int(a)
        d = v + 1 if v + 1 != 0 else v + 2             # never use 0 as a foil
        if len(str(abs(d))) != len(str(abs(v))):        # keep digit count stable
            d = v - 1 if v - 1 != 0 else v - 2
        return str(d)

    num, den = re.fullmatch(r'(-?\d+)/(\d+)', a).groups()   # simple fractions
    return '%d/%s' % (int(num) + 1, den)
\end{lstlisting}

\subsection{Noise scale as the descent threshold}

Because $\phi$ is estimated by a finite forward pass of a scorer model, each
increment $\Delta\phi_k=\phi_k-\phi_{k-1}$ carries estimation jitter on
top of genuine progress, so a threshold is needed before a negative increment may
be called a descent. Writing the observed potential as signal plus independent
noise, $\phi_k=s_k+\eta_k$ with $\eta_k$ i.i.d.\ of variance $\sigma^2$, first
differencing turns the noise into the MA(1) term $\eta_k-\eta_{k-1}$, whose lag-1
autocovariance is exactly $-\sigma^2$, while a smooth signal drift contributes
approximately zero at lag~1:
\[
  \mathrm{Cov}\bigl(\Delta\phi_k,\Delta\phi_{k+1}\bigr)\;\approx\;-\sigma^2 .
\]
A negative lag-1 autocovariance is therefore the fingerprint of the noise alone,
and we take
$\sigma=\sqrt{-\mathrm{Cov}(\Delta\phi_k,\Delta\phi_{k+1})}$ as the noise
scale, which serves directly as the descent threshold: a step counts as a
significant descent only when $\Delta\phi_k<-\sigma$, whereas a step with
$|\Delta\phi_k|\le\sigma$ is indistinguishable from noise and is left to be
learned, neither breaking a run nor counting toward its length. Since $\sigma$
shares the units of $\Delta\phi$, the criterion introduces no tunable
threshold and adapts to each trace.

\begin{lstlisting}
import numpy as np

def noise_scale(phi):
    '''sigma from the lag-1 autocovariance of the increments:
       Cov(d_k, d_{k+1}) = -sigma^2 for i.i.d. noise.'''
    d  = np.diff(np.asarray(phi, dtype=np.float64))
    x  = d - d.mean()
    c1 = float(np.mean(x[:-1] * x[1:]))            # lag-1 autocovariance
    return float(np.sqrt(max(0.0, -c1)))

# the noise scale is the threshold itself
# significant descent:  d < -sigma
# noise-level step (|d| <= sigma): learned, does not break or extend a run
\end{lstlisting}

\section{Why REP Provides a Robust Signal of Reasoning Progress}
\label{app:rep_analysis}

We provide a formal interpretation of REP from the perspective of perplexity and show why its change provides a more robust directional signal than gold-answer perplexity alone.

\paragraph{REP as Relative Perplexity.}
For a candidate answer \(a\), define its length-normalized negative log-likelihood at reasoning step \(k\) as
\begin{equation}
\mathcal{N}_k^{a}
=
-\frac{1}{L}
\sum_{j=1}^{L}
\log p_M
\left(
a_j\mid q,S_{\leq k},r,a_{<j}
\right).
\end{equation}
By definition,
\(\mathrm{PPL}_k^{a}=\exp(\mathcal{N}_k^{a})\).
Therefore, Eq.~\ref{eq:rep} can be equivalently written as
\begin{equation}
\phi_k
=
\mathcal{N}_k^{\mathrm{foil}}
-
\mathcal{N}_k^{\mathrm{gold}}
=
\log
\frac{
\mathrm{PPL}_k^{\mathrm{foil}}
}{
\mathrm{PPL}_k^{\mathrm{gold}}
}.
\label{eq:rep_ppl_proof}
\end{equation}
Hence, REP measures the perplexity of the foil relative to the gold answer. A larger \(\phi_k\) means that the scoring model finds the gold answer relatively easier to predict than its matched foil.

\paragraph{Robustness to Common-Mode Likelihood Fluctuations.}
Consider a common-mode perturbation that affects both candidates under the same reasoning prefix. We decompose their observed log perplexities as
\begin{equation}
\log \mathrm{PPL}_k^{a}
=
\bar{\mathcal{N}}_k^{a}
+
c_k
+
\epsilon_k^{a},
\qquad
a\in\{\mathrm{gold},\mathrm{foil}\},
\label{eq:common_mode}
\end{equation}
where \(\bar{\mathcal{N}}_k^{a}\) denotes the candidate-specific signal of interest, \(c_k\) is a prefix-dependent fluctuation shared by both candidates, and \(\epsilon_k^{a}\) is candidate-specific residual noise.

Using gold-answer PPL alone gives
\begin{equation}
\log \mathrm{PPL}_k^{\mathrm{gold}}
=
\bar{\mathcal{N}}_k^{\mathrm{gold}}
+
c_k
+
\epsilon_k^{\mathrm{gold}},
\end{equation}
so its variation is directly contaminated by the common term \(c_k\). In contrast, substituting Eq.~\ref{eq:common_mode} into Eq.~\ref{eq:rep_ppl_proof} yields
\begin{equation}
\phi_k
=
\bar{\mathcal{N}}_k^{\mathrm{foil}}
-
\bar{\mathcal{N}}_k^{\mathrm{gold}}
+
\epsilon_k^{\mathrm{foil}}
-
\epsilon_k^{\mathrm{gold}},
\end{equation}
where the shared fluctuation \(c_k\) cancels exactly. Therefore, when likelihood variation is dominated by common effects shared by the matched gold--foil pair, REP isolates their relative preference more faithfully than either candidate's absolute PPL.

The matched construction of the foil further supports this assumption: the gold and foil share the same reasoning prefix, answer format, and token length, differing only by a minimal answer perturbation. Their nuisance likelihood fluctuations are therefore expected to be strongly correlated, making contrastive cancellation particularly effective.

\paragraph{Directional Meaning of \(\Delta\phi_k\).}
To make the meaning of REP change explicit, define the perplexity improvement of a candidate answer \(a\) after incorporating reasoning step \(S_k\) as
\begin{equation}
I_k^{a}
=
\log
\frac{\mathrm{PPL}_{k-1}^{a}}
{\mathrm{PPL}_{k}^{a}},
\qquad
a\in\{\mathrm{gold},\mathrm{foil}\}.
\end{equation}
A positive \(I_k^{a}\) indicates that the candidate becomes easier to predict after adding \(S_k\). From Eq.~\ref{eq:rep},
\begin{align}
\Delta\phi_k
&=
\phi_k-\phi_{k-1} \notag\\
&=
\log
\frac{\mathrm{PPL}^{\mathrm{foil}}_k}
{\mathrm{PPL}^{\mathrm{gold}}_k}
-
\log
\frac{\mathrm{PPL}^{\mathrm{foil}}_{k-1}}
{\mathrm{PPL}^{\mathrm{gold}}_{k-1}}
\notag\\
&=
\log
\frac{\mathrm{PPL}^{\mathrm{gold}}_{k-1}}
{\mathrm{PPL}^{\mathrm{gold}}_k}
-
\log
\frac{\mathrm{PPL}^{\mathrm{foil}}_{k-1}}
{\mathrm{PPL}^{\mathrm{foil}}_k}
\notag\\
&=
I_k^{\mathrm{gold}}
-
I_k^{\mathrm{foil}}.
\label{eq:delta_rep_improvement}
\end{align}
Therefore, \(\Delta\phi_k>0\) means that the gold answer improves in predictability more than its matched foil, while \(\Delta\phi_k<0\) means that the foil improves relatively more. Importantly, a positive \(\Delta\phi_k\) does not require the foil perplexity to increase; both candidates may become easier to predict, as long as the gold answer improves by a larger relative margin. Thus, \(\Delta\phi_k\) directly measures whether reasoning step \(S_k\) strengthens or weakens the model's relative support for the target answer.

\section{Prompt for Gemini-based Segment Annotation}
\label{app:annotation_prompt}

We use the following prompt to instruct Gemini-3.6-Flash to annotate whether each reasoning step should contribute to SFT supervision.

\begin{tcolorbox}[
    colback=gray!3,
    colframe=gray!45,
    boxrule=0.6pt,
    arc=2pt,
    left=5pt,
    right=5pt,
    top=5pt,
    bottom=5pt,
    breakable
]
\footnotesize
\ttfamily
You are helping curate training data for supervised fine-tuning (SFT) of a smaller student reasoning model. You are given a math PROBLEM, its correct FINAL ANSWER, and a reasoning trace that has been split into indexed SEGMENTS. For EACH segment decide whether it is NECESSARY for the student to learn.

label 1 = NECESSARY: it carries an essential logical step, key insight, correct setup, or computation that genuinely advances the derivation toward the answer.

label 0 = UNNECESSARY: it is redundant, repetitive, filler, rambling self-talk, hesitation, backtracking that leads nowhere, restating known facts, or an ERRONEOUS step that does not help (or could mislead) the final correct solution.

Judge each segment on its marginal contribution; a fluent trace can still contain many skippable segments.

\vspace{0.5em}
=== PROBLEM ===

<question>

\vspace{0.5em}
=== CORRECT FINAL ANSWER ===

<answer>

\vspace{0.5em}
=== REASONING SEGMENTS (indices 0..N-1) ===

[SEGMENT 0]

<segment 0 text>

[SEGMENT 1]

<segment 1 text>

...

[SEGMENT N-1]

<segment N-1 text>

\vspace{0.5em}
=== TASK ===

Label all N segments (indices 0..N-1).

OUTPUT REQUIREMENT: Return ONLY a single JSON object, no prose, no markdown fences, with EXACTLY two keys:

"labels": an object mapping every segment index (as string) to 0 or 1.

"reasons": an object mapping every segment index (as string) to a concise (<=25 words) English reason.

You MUST include an entry for every segment index shown.

Example:

\{"labels": \{"0": 1, "1": 0\}, "reasons": \{"0": "sets up the key substitution", "1": "restates prior step, redundant"\}\}
\end{tcolorbox}

\section{Case Study: Original Problem and Full Reasoning Trajectory}
\label{app:case_study_limo539}

This appendix provides the original problem and the complete long-form reasoning trajectory of LIMO \#539 used in our case study. We preserve the original reasoning content verbatim and organize it according to the same reasoning-step segmentation used in our analysis.

\subsection{Original Problem}

\begin{lstlisting}[style=reasoningtrace]
There is a set of 1000 switches, each of which has four positions, called $A, B, C$ , and $D$ .  When the position of any switch changes, it is only from $A$ to $B$ , from $B$ to $C$ , from $C$ to $D$ , or from $D$ to $A$ .  Initially each switch is in position $A$ .  The switches are labeled with the 1000 different integers $(2^{x})(3^{y})(5^{z})$ , where $x, y$ , and $z$ take on the values $0, 1, \ldots, 9$ .  At step $i$ of a 1000-step process, the $i$ -th switch is advanced one step, and so are all the other switches whose labels divide the label on the $i$ -th switch.  After step 1000 has been completed, how many switches will be in position $A$ ?
\end{lstlisting}

\paragraph{Gold Answer.} 650

\subsection{Full Reasoning Trajectory}

\paragraph{Reasoning Step 0.}
\begin{lstlisting}[style=reasoningtrace]
Okay, let's try to tackle this problem. So, we have 1000 switches, each labeled with a number of the form 2^x * 3^y * 5^z where x, y, z range from 0 to 9. Each switch starts in position A, and every time a switch is toggled, it cycles through A→B→C→D→A and so on. 

The process is a 1000-step process where in step i, we advance the i-th switch and all switches with labels that divide the label of the i-th switch. After completing all 1000 steps, we need to find how many switches are back in position A.

First, let's make sure I understand the setup correctly. Each switch is uniquely identified by (x, y, z) where each of x, y, z is between 0 and 9 inclusive. The label is 2^x * 3^y * 5^z. So, the number of switches is 10*10*10 = 1000, which checks out.

In each step i, we toggle the i-th switch and all switches with labels that divide the i-th switch's label. So, for example, if the i-th switch has label 2^3 * 3^2 * 5^1, then any switch with label 2^a * 3^b * 5^c where a ≤ 3, b ≤ 2, c ≤ 1 will be toggled. Each toggle moves the switch one position forward (A→B, B→C, etc.), and after four toggles, it cycles back to A.

Therefore, the final position of a switch depends on how many times it was toggled modulo 4. If it was toggled 0 times mod 4, it's in A; 1 mod 4, it's in B; etc. So, switches that are toggled a multiple of 4 times will be back in A.

Hence, our problem reduces to: For each switch, count the number of multiples of its label in the list of 1000 labels, then compute how many switches have this count ≡ 0 mod 4.
\end{lstlisting}

\paragraph{Reasoning Step 1.}
\begin{lstlisting}[style=reasoningtrace]


Wait, actually, not quite. Because in each step i, we toggle all switches whose labels divide the label of switch i. So, for a given switch j, the number of times it is toggled is equal to the number of switches i such that the label of j divides the label of i. In other words, the number of multiples of label_j in the set of labels.

Therefore, for each switch j, we need to count how many labels in the set are multiples of label_j, and then see if that number is divisible by 4. If yes, then switch j is back in A.

Therefore, the problem is equivalent to: For each label in the set, compute the number of its multiples in the entire set, then count how many labels have this number ≡ 0 mod 4.

So, first step is to compute, for each switch j (with label L_j), the number of switches i (with label L_i) such that L_j divides L_i. Let's denote this count as C_j. Then, we need the number of j for which C_j ≡ 0 mod 4.

So, how do we compute C_j for each label L_j?

Given that each label is of the form 2^x * 3^y * 5^z, where x, y, z ∈ {0,1,...,9}. So, for a given label L_j = 2^a * 3^b * 5^c, the number of multiples of L_j in the set is equal to the number of labels L_i = 2^x * 3^y * 5^z such that x ≥ a, y ≥ b, z ≥ c. Since each exponent in L_i must be at least the corresponding exponent in L_j.

Therefore, for each switch j with exponents (a, b, c), the number of multiples C_j is (10 - a) * (10 - b) * (10 - c). Because for each exponent x, y, z, they can range from a to 9, inclusive. So, the number of possible exponents for 2 is 10 - a (since x can be a, a+1, ..., 9), similarly for y and z. Hence, multiplying them gives the total number of multiples.

Therefore, C_j = (10 - a) * (10 - b) * (10 - c), where (a, b, c) are the exponents of the prime factors of L_j.

Therefore, we need to compute for each triple (a, b, c), the value (10 - a)(10 - b)(10 - c), then check if that number is divisible by 4. Then, count the number of triples (a, b, c) where this is true.

So, the problem reduces to counting the number of triples (a, b, c) where (10 - a)(10 - b)(10 - c) ≡ 0 mod 4.

But actually, since (10 - a) can range from 1 (when a=9) to 10 (when a=0), same for (10 - b) and (10 - c). So, each of these terms is an integer between 1 and 10. Therefore, the product (10 - a)(10 - b)(10 - c) is a number between 1 and 1000.

We need to find the number of triples (a, b, c) such that (10 - a)(10 - b)(10 - c) is divisible by 4, i.e., ≡ 0 mod 4.
\end{lstlisting}

\paragraph{Reasoning Step 2.}
\begin{lstlisting}[style=reasoningtrace]


Wait, but actually, no. Because the count C_j = (10 - a)(10 - b)(10 - c). We need C_j ≡ 0 mod 4. So, we need the product (10 - a)(10 - b)(10 - c) ≡ 0 mod 4.

But to clarify, (10 - a)(10 - b)(10 - c) must be congruent to 0 modulo 4. So, the product must be divisible by 4.

Therefore, for each triple (a, b, c), compute (10 - a)(10 - b)(10 - c) and check if it's divisible by 4. The total number of such triples is the answer.

So, the problem becomes: How many triples (a, b, c) with a, b, c ∈ {0,1,...,9} are such that (10 - a)(10 - b)(10 - c) is divisible by 4?
\end{lstlisting}

\paragraph{Reasoning Step 3.}
\begin{lstlisting}[style=reasoningtrace]


Alternatively, since a, b, c ∈ {0,...,9}, 10 - a ∈ {1,...,10}, same for 10 - b and 10 - c.

Therefore, let's let x = 10 - a, y = 10 - b, z = 10 - c, so x, y, z ∈ {1, 2, ..., 10}. Then, we need to count the number of triples (x, y, z) where x, y, z ∈ {1, ..., 10} and xyz ≡ 0 mod 4.

But the original problem counts each (a, b, c) once, so since x = 10 - a, a is determined by x: a = 10 - x. Similarly for y and z. Therefore, each (x, y, z) corresponds to exactly one (a, b, c). So, the number of such triples is the same as the number of (x, y, z) with x, y, z ∈ {1, ..., 10} and xyz divisible by 4.

Therefore, our problem is equivalent to: How many triples (x, y, z) with x, y, z ∈ {1, 2, ..., 10} have xyz divisible by 4?

So, the answer will be the number of such triples, which is the same as the number of switches that are toggled a multiple of 4 times, hence ending up back in position A.

So, the problem simplifies to: How many triples (x, y, z) with each component from 1 to 10 inclusive have their product divisible by 4?

To solve this, perhaps it's easier to compute the total number of triples (10*10*10 = 1000) and subtract the number of triples where the product is not divisible by 4. So, total - number of triples with product ≡ 1, 2, or 3 mod 4. But since 4 divides the product if and only if the product has at least two factors of 2. So, equivalently, the product is divisible by 4 iff the product has at least two 2s in its prime factorization.

Therefore, we can think in terms of the number of triples (x, y, z) where the exponents of 2 in x, y, z sum to at least 2. But perhaps another approach is better.
\end{lstlisting}

\paragraph{Reasoning Step 4.}
\begin{lstlisting}[style=reasoningtrace]


Alternatively, note that in the numbers 1 to 10, the even numbers are 2, 4, 6, 8, 10. The odd numbers are 1, 3, 5, 7, 9. Let's categorize each number as:

- Odd: 1, 3, 5, 7, 9 (numbers not divisible by 2)
- Even but not divisible by 4: 2, 6, 10 (numbers divisible by 2 but not by 4)
- Divisible by 4: 4, 8 (numbers divisible by 4)

Let me check:

- 1: odd
- 2: even, not divisible by 4
- 3: odd
- 4: divisible by 4
- 5: odd
- 6: even, not divisible by 4
- 7: odd
- 8: divisible by 4
- 9: odd
- 10: even, not divisible by 4

So, counts:

- Odd numbers (O): 5
- Even but not divisible by 4 (E): 3 (numbers 2,6,10)
- Divisible by 4 (F): 2 (numbers 4,8)

Therefore, for each x, y, z, they can be O, E, or F.

We need to compute the number of triples (x, y, z) where the product xyz is divisible by 4. Let's think in terms of the contributions from each component.

The product xyz will have a certain number of factors of 2. For the product to be divisible by 4, the total number of factors of 2 must be at least 2.

Each O contributes 0 factors of 2.

Each E contributes 1 factor of 2 (since they are divisible by 2 but not 4).

Each F contributes 2 factors of 2 (since they are divisible by 4).

Therefore, the total number of factors of 2 in the product is:

If a component is O: 0

If E: 1

If F: 2

So, for each triple (X, Y, Z), where each is O, E, or F, the total number of factors of 2 is:

Total = factors from X + factors from Y + factors from Z.

We need Total ≥ 2.

Therefore, the number of triples where Total ≥ 2 is equal to the total number of triples minus the number of triples where Total < 2, which is Total = 0 or 1.

Total = 0: all three components are O. So, O, O, O.

Total = 1: exactly one component is E or F contributing 1 or 2 factors, but total factors sum to 1. But wait, if a component is F, it contributes 2 factors, which would already make Total ≥ 2. So, to get Total = 1, we need exactly one of the components to be E (contributing 1 factor) and the rest to be O. Because if any component is F, it would add 2, which would overshoot.

Therefore, Total = 1 is only possible when exactly one component is E and the other two are O.

Therefore, the number of triples where Total < 2 is:

Number of O,O,O triples + number of triples with exactly one E and two O's.

Compute these:

Number of O,O,O triples: since there are 5 O's in each position. So, 5*5*5 = 125.

Number of triples with exactly one E and two O's: For each position (x, y, z), we can have E in x, O in y and z; E in y, O in x and z; or E in z, O in x and y. For each case, the number is (number of E choices) * (number of O choices)^2. There are 3 E's, 5 O's. So, for each single position being E, it's 3 * 5 * 5. Since there are 3 positions, total is 3 * 3 * 5 * 5 = 225.

Therefore, total number of triples with Total < 2 is 125 + 225 = 350.

Therefore, the number of triples with Total ≥ 2 is total triples - 350 = 1000 - 350 = 650.
\end{lstlisting}

\paragraph{Reasoning Step 5.}
\begin{lstlisting}[style=reasoningtrace]


Wait, but hold on. Wait, total triples are 10*10*10 = 1000. Wait, no, in our transformed problem, x, y, z are from 1 to 10, which corresponds to the original 1000 switches. Therefore, the total number of triples is indeed 1000. So, 1000 - 350 = 650. Therefore, 650 switches are toggled a number of times divisible by 4, hence ending in position A.
\end{lstlisting}

\paragraph{Reasoning Step 6.}
\begin{lstlisting}[style=reasoningtrace]


But wait, hold on. Wait, no. Wait, in the transformed problem, we have x, y, z each from 1 to 10, corresponding to (10 - a), (10 - b), (10 - c). So, the count of triples (x, y, z) where xyz is divisible by 4 is 650, which would mean 650 switches are toggled a number of times ≡ 0 mod 4, hence in position A.

But the answer is 650? Let me verify.
\end{lstlisting}

\paragraph{Reasoning Step 7.}
\begin{lstlisting}[style=reasoningtrace]


Wait, let's check if the logic is correct.

First, in the transformed variables, we have x, y, z from 1 to 10, and we need xyz divisible by 4. So, we categorize numbers 1-10 into O, E, F as before.

Total triples: 1000.

Number of bad triples (Total < 2 factors of 2):

- All O: 5*5*5 = 125

- Exactly one E and two O's: 3 positions * 3 E numbers * 5 O * 5 O = 3*3*25 = 225

Total bad: 125 + 225 = 350

Thus, good triples: 1000 - 350 = 650

Therefore, the answer is 650.

But let me think again. Wait, but hold on. When we have a component that is F (divisible by 4), it contributes 2 factors of 2. So, if there's at least one F in the triple, then the total factors of 2 is at least 2. Similarly, if there are two E's, since each E contributes 1 factor of 2, two E's would give 2 factors. Similarly, one E and one F would give 3 factors, etc.

Therefore, the number of triples with product divisible by 4 is:

- All triples with at least one F: Let's compute this.

Number of triples with at least one F: total triples - triples with no F.

Triples with no F: all components are either O or E. There are (5 + 3) = 8 choices per component. So, 8*8*8 = 512. But within these, some may still have products divisible by 4. Wait, no. Wait, if there are no Fs, then all components are either O or E. So, the factors of 2 come only from E's, which contribute 1 each. So, to get the product divisible by 4, we need at least two E's. So, in the absence of F's, the number of triples with product divisible by 4 is the number of triples with at least two E's.
\end{lstlisting}

\paragraph{Reasoning Step 8.}
\begin{lstlisting}[style=reasoningtrace]


Wait, this seems conflicting with the previous approach. Let me check.
\end{lstlisting}

\paragraph{Reasoning Step 9.}
\begin{lstlisting}[style=reasoningtrace]


Alternatively, maybe I should use inclusion-exclusion properly.

Total number of triples with product divisible by 4 = number of triples with at least two factors of 2.

Which can happen in two distinct cases:

1. At least one F (each F gives two factors), so even if the rest are O, the total factors would be at least 2.

2. No F's, but at least two E's (each E gives one factor, so two E's give two factors).

Therefore, total number of good triples = (number of triples with at least one F) + (number of triples with no F's and at least two E's).

Compute each part.

First, number of triples with at least one F: total triples - triples with no F's. As before, triples with no F's: each component can be O or E, 8 choices per component. So 8^3 = 512. Therefore, triples with at least one F = 1000 - 512 = 488.

Second, number of triples with no F's and at least two E's: Within the 512 triples with no F's, how many have at least two E's.

In triples with no F's, each component is O or E.

Number of such triples with at least two E's = total triples with no F's - triples with fewer than two E's.

Which is 512 - [triples with zero E's + triples with one E].

Number of triples with zero E's (all O's): 5^3 = 125.

Number of triples with one E: 3 positions * 3 E's * 5^2 O's = 3*3*25 = 225.

Therefore, triples with at least two E's and no F's: 512 - 125 - 225 = 512 - 350 = 162.

Therefore, total good triples = 488 (with at least one F) + 162 (with no F's but two E's) = 650.

Same as before. Therefore, 650 is the correct count. Therefore, the answer is 650.
\end{lstlisting}

\paragraph{Reasoning Step 10.}
\begin{lstlisting}[style=reasoningtrace]


But wait, let me cross-verify with another method.

Alternative approach: For each triple (x, y, z), the product xyz is divisible by 4 iff:

Either:

- At least two of x, y, z are even, or

- At least one of x, y, z is divisible by 4.
\end{lstlisting}

\paragraph{Reasoning Step 11.}
\begin{lstlisting}[style=reasoningtrace]


Wait, actually, not exactly. Let's think:

A number is divisible by 4 if:

- It has at least two factors of 2. So, in the product xyz, the total number of factors of 2 must be at least two.

Each even number contributes at least one factor of 2, and numbers divisible by 4 contribute two.

So, the product's number of factors of 2 is the sum over each component's factors of 2.

Therefore, possible scenarios:

1. At least one component is divisible by 4 (i.e., contributes 2 factors), and the rest can be anything. Then, total factors is at least 2.

2. No components are divisible by 4, but at least two components are even (i.e., each contributes 1 factor), totaling 2.

Therefore, total good triples are:

Number of triples with at least one divisible by 4 + number of triples with no divisible by 4 but at least two even.

Compute each:

First, number of triples with at least one divisible by 4: as before, 1000 - 512 = 488.

Second, number of triples with no divisible by 4 but at least two even: In this case, all components are either O or E (not F). So, components can be O or E.

Number of such triples: 8^3 = 512 (as before).

Within these, number of triples with at least two E's: which is 162, as before.

Therefore, total is 488 + 162 = 650. Same result.
\end{lstlisting}

\paragraph{Reasoning Step 12.}
\begin{lstlisting}[style=reasoningtrace]


Alternatively, perhaps even another way. Let's compute the number of triples where product is not divisible by 4, i.e., product is 1 or 2 mod 4.

So, product is odd (1 mod 4) or product is 2 mod 4.

Product is odd: all components are odd. Number of such triples: 5^3 = 125.

Product is 2 mod 4: exactly one component is 2 mod 4 and the rest are odd. But wait, in our numbers, E is numbers divisible by 2 but not 4, i.e., 2 mod 4 or 6 mod 4 (which is 2 mod 4) or 10 mod 4 (which is 2 mod 4). So, E numbers are 2, 6, 10, which are all 2 mod 4.

Therefore, if exactly one component is E (2 mod 4) and the rest are O (1 mod 4), then the product is 2*1*1 = 2 mod 4. So, number of such triples is 3 positions * 3 E's * 5 O's * 5 O's = 3*3*5*5 = 225.

Therefore, total bad triples: 125 + 225 = 350. Hence, good triples: 1000 - 350 = 650. Same answer.

Therefore, regardless of the approach, we get 650. So, the answer should be 650.

But let me verify with an example.

Take a smaller case. Suppose we have numbers from 1 to 2 instead of 1 to 10, with similar rules. Let's see if the formula works.

But maybe that's overcomplicating. Alternatively, check if the counts make sense.

Given that 5 numbers are odd, 3 are E (2 mod 4), and 2 are F (0 mod 4). So, total 10 numbers.

Compute the number of triples (x, y, z) with product divisible by 4.

Using the formula, total triples 1000, bad triples 350, good triples 650. Hmm. Alternatively, perhaps we can compute the exact number.
\end{lstlisting}

\paragraph{Reasoning Step 13.}
\begin{lstlisting}[style=reasoningtrace]


Alternatively, think of the multiplicative contributions.

For each position x, y, z:

- The probability that x is O is 5/10 = 1/2.

- The probability that x is E is 3/10.

- The probability that x is F is 2/10 = 1/5.

But since we are counting exact numbers, not probabilities, but maybe using generating functions.

But perhaps overkill. Since all methods so far point to 650, and the logic seems consistent, I think 650 is the correct answer.

But to be thorough, let's check a specific case.

Take a switch with label 1 (i.e., x=y=z=0). Then, the number of multiples is (10 - 0)*(10 - 0)*(10 - 0) = 1000. So, 1000 toggles. 1000 mod 4 is 0, so this switch would be in A. But according to our count, does (x, y, z) = (10, 10, 10) correspond to (a, b, c) = (0, 0, 0). Then, the product xyz = 10*10*10 = 1000. 1000 mod 4 is 0, so it's counted as a good triple. Therefore, this switch is included in the 650.

Similarly, take a switch with label 2^9 * 3^9 * 5^9. Its number of multiples is (10 - 9)*(10 - 9)*(10 - 9) = 1*1*1 = 1. So, 1 toggle. 1 mod 4 is 1, so it's in position B. Therefore, not in A. So, this switch is excluded from the 650, which is correct.
\end{lstlisting}

\paragraph{Reasoning Step 14.}
\begin{lstlisting}[style=reasoningtrace]


Another test case: take a switch with label 2^8 (i.e., x=8, y=0, z=0). Then, the number of multiples is (10 - 8)*(10 - 0)*(10 - 0) = 2*10*10 = 200. 200 mod 4 is 0, so this switch would be in A. Indeed, in our transformed variables, x = 10 - a = 2, y = 10 - b = 10, z = 10 - c = 10. The product xyz = 2*10*10 = 200. 200 is divisible by 4, so this triple is counted in the 650.
\end{lstlisting}

\paragraph{Reasoning Step 15.}
\begin{lstlisting}[style=reasoningtrace]


Another test case: take a switch with label 2^1 * 3^0 * 5^0 (i.e., x=1, y=0, z=0). Then, the number of multiples is (10 - 1)*(10 - 0)*(10 - 0) = 9*10*10 = 900. 900 mod 4 is 900 / 4 = 225, remainder 0. So, 900 mod 4 = 0. Therefore, this switch is in A. In transformed variables, x = 9, y = 10, z = 10. The product is 9*10*10 = 900, which is divisible by 4. Wait, 9 is odd, 10 is even. 9*10*10 = 9*100 = 900. 900 / 4 = 225. Yes, divisible by 4. So, even though x is odd (9), the product is still divisible by 4 because y and z are both even. So, this is accounted for in our count. Since y and z are 10, which are E numbers (divisible by 2 but not 4). So, each contributes 1 factor of 2. Therefore, total factors of 2 is 2, so product is divisible by 4. Hence, included in the 650.

Therefore, all test cases check out.

Therefore, I'm confident that the answer is 650.

**Final Answer**
\boxed{650}
\end{lstlisting}

\end{document}